%% file: main.tex
\documentclass[journal]{IEEEtran}

\usepackage{cite}
\usepackage{amsmath,amssymb,amsfonts}
\usepackage{graphicx}
\usepackage{xcolor}
\usepackage{booktabs}
\usepackage{array}
\usepackage{url}
\usepackage{hyperref}

\hypersetup{hidelinks}
\usepackage{amsmath}
\usepackage{bm}

\begin{document}

\title{TISC: A Text-Driven Image Semantic Communication System for Faithful Reconstruction}

\author{
Feifan~Zhang$^*$, 
Yuyang~Du$^*$, 
Xiaoyan~Liu, 
Soung~Chang~Liew~\IEEEmembership{Life Fellow,~IEEE}

\thanks{F. Zhang, Y. Du, X. Liu and S. C. Liew are with the Department of Information Engineering, The Chinese University of Hong Kong, HKSAR, China. S. C. Liew is the corresponding author (e-mail:soung@ie.cuhk.edu.hk).}
\thanks{The work was supported in part by the Hong Kong Innovation and Technology Fund (Project Number: ITS/362/24), and in part by the Shenzhen-Hong Kong-Macao Science and Technology Program (STIC), Category C (Project Number: SGDX20230821094359004). The experimental work in this paper was conducted in the JC STEM Lab of Advanced Wireless Networks for Mission-Critical Automation and Intelligence funded by The Hong Kong Jockey Club Charities Trust}
\thanks{$^*$F. Zhang and Y. Du contribute equally to this work.}
\vspace{-1.5em}
}

\maketitle

\begin{abstract}
Generative image semantic communication converts an image into a text description and then performs text-to-image reconstruction at the receiver via diffusion-based generative models. This paradigm has attracted broad attention due to its extremely low bandwidth cost. However, existing methods still face two critical bottlenecks across image-to-text (I2T) semantic extraction at the transmitter and text-to-image (T2I) semantic reconstruction at the receiver: (i) semantic loss and distortion in I2T, where holistic image descriptions may omit fine-grained object attributes and spatial-position information, causing the generated text to deviate from the original image semantics; and (ii) insufficient semantic faithfulness in T2I, where even with the same semantically faithful text description, different initial noise settings may lead diffusion-based reconstruction to produce images with different levels of semantic consistency with the original image. These issues jointly limit the semantic faithfulness of image reconstruction. To address them, we propose TISC, a text-driven image semantic communication framework tailored for faithful reconstruction. TISC incorporates two key designs: (1) Tree-Structured Attribute Semantic Extraction (TSASE), which decomposes semantic extraction into global scene, background, and object-level attribute descriptions, covering spatial position, shape/pose, color, material, and other physical attributes for each detected object; and (2) an Initial Noise Optimization (INO) mechanism, which selects an initial noise seed at the transmitter according to a comprehensive similarity score that jointly considers visual and semantic consistency. Experiments on multiple datasets show that TSASE improves object-position recovery and semantic description faithfulness, while the INO parameter study supports the adopted configuration for noise selection.
\end{abstract}


\input{sections/01_introduction}
\input{sections/02_tsase}
\input{sections/03_ino}
\input{sections/04_experiments}
\input{sections/05_conclusion}
\appendices
\input{sections/appendix_a}

\bibliographystyle{IEEEtran}
\bibliography{references}

\end{document}

%% file: sections/01_introduction.tex
\section{Introduction}
\label{sec:introduction}

With the evolution from 1G to 5G, data transmission rates in wireless communication systems have continued to increase\cite{5g}. However, cutting-edge physical-layer designs are now approaching the Shannon limit, making further improvements in effective throughput increasingly challenging\cite{shannon,beyondshann}. To meet the critical bandwidth requirement for future 6G, research in wireless networks is now moving beyond traditional architectures to semantic communication systems \cite{semantic1,semantic2,semantic3}.

A semantic communication system captures the essential properties of information that are relevant to specific applications. By generating, processing, and transmitting only the most critical information, it reduces channel resource usage while maintaining effectiveness for the target applications. Semantic communications have been widely investigated as a key enabler for next-generation wireless networks \cite{semantic4,semantic5,semantic6}.

Image transmission is a major application area of semantic communication. Classic image semantic communication systems send compressed visual features, such as semantic segmentation maps \cite{OOSC,semanticmap2,semanticmap3}, latent representations \cite{latent1,Deepjscc,NTSCC,saoosc,transformerjscc,swinjscc,deepjscc-f}, or hierarchical image features \cite{semanticmap4}, over the communication link and recover the image at the receiver side with a well trained AI model tailored to feature-based image reconstruction. Recent technical advancements in multimodal large language models (MLLMs) open up a new possibility for image semantic communications \cite{mllmsurvey}: the semantic information of an image can be summarized into text descriptions by a transmitter-side MLLM for transmission to the receiver. The receiver then uses a generative model to convert the text back to an image. This emerging paradigm, also known as text-driven image semantic communication, introduces significantly reduced bandwidth usage compared to conventional pixel-wise image transmission or feature-based image semantic communication. We further verify this bandwidth advantage in the subsequent experiment. The results show that, at a comparable level of reconstruction similarity, the proposed text-driven semantic transmission scheme requires over 20 times fewer compressed bits than conventional image transmission (see Appendix~\ref{sec:appendix} for details). This finding further demonstrates the potential of text-driven image semantic communication for low-bandwidth image transmission.

\begin{figure*}[t]
    \centering
    \includegraphics[width=0.98\textwidth]{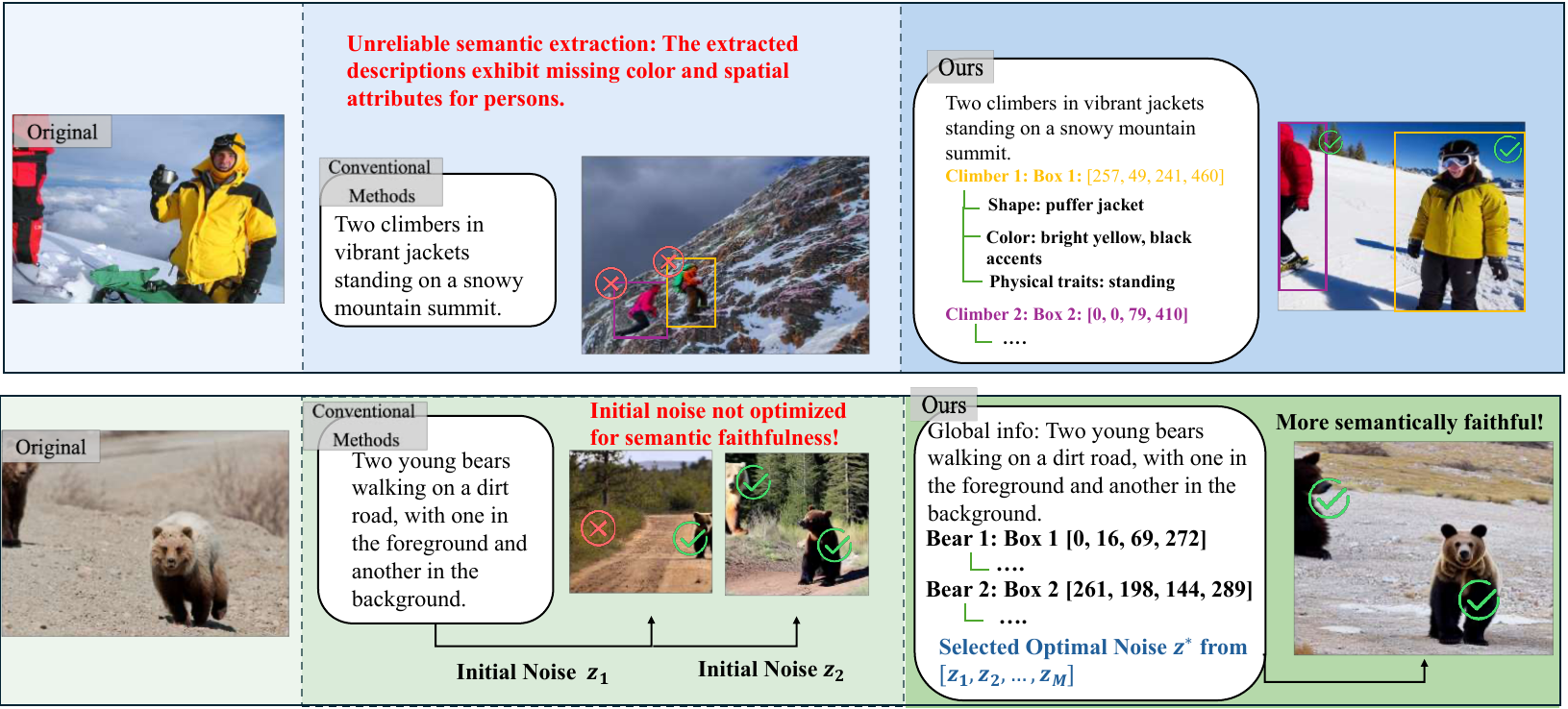}
    \caption{Comparative illustrations of conventional methods versus our generative semantic communication framework from two aspects: (a) I2T semantic distortion: conventional methods may suffer from attribute-level semantic distortions such as missing or incorrect color and spatial information, which impair semantic extraction, while our approach mitigates such distortions and improves the reliability of semantic extraction; (b) insufficient semantic faithfulness in T2I reconstruction: conventional methods randomly sample the initial noise, so different noises may lead to reconstruction results with different semantic faithfulness even under the same text description. The proposed method evaluates multiple candidate initial noises at the transmitter and selects the one that produces the most faithful reconstruction within the candidate set.}
    \label{fig:comparison}
\end{figure*}

Despite pioneering efforts in this line of research \cite{txt_sem1,txt_sem2,txt_sem3,addressOOD}, existing works exhibit notable limitations in semantic faithfulness in both I2T semantic extraction at the transmitter side and T2I semantic reconstruction at the receiver side. Fig.~\ref{fig:comparison} illustrates two existing problems.

\textbf{Incomplete and distorted I2T semantic representation:} The MLLM for I2T semantic extraction may produce image descriptions that omit critical details from the original image or inaccurately capture certain semantic elements \cite{MLLM_hallucination}. Such issues can arise in the representation of the number of objects, visual attributes such as colors and materials, or spatial attributes such as object positions and relationships \cite{hallucination2}. When these text descriptions are used for T2I at the receiver, the resulting image may exhibit inconsistencies in object attributes or errors in spatial layouts.

\textbf{Insufficient semantic faithfulness in T2I reconstruction:} Even if the transmitter extracts and transmits a semantically faithful text description, the receiver may still fail to reconstruct an image that is semantically consistent with the original one. This is because diffusion-based generation starts from a random initial noise, and different initial noises may guide the generation process along different trajectories \cite{diffusion}. Therefore, even with the same text description, different initial noises can still lead to reconstructed images with varying degrees of semantic fidelity \cite{noise}.

Overall, we need to answer two outstanding questions.

\textbf{Question 1:} \textit{I2T -- How can we extract a semantically faithful text description from the original image, thereby reducing information loss and distortion during semantic extraction?}

\textbf{Question 2:} \textit{T2I -- Given a semantically faithful text description, how can we generate a reconstructed image that is semantically consistent with the original image?}

This paper puts forth TISC, a text-driven image semantic communication framework addressing these questions through the following technical innovations. First, as illustrated in Fig.~\ref{fig:comparison}a, we develop a Tree-Structured Attribute Semantic Extraction (TSASE) module to realize accurate semantic representations. This approach decomposes complex semantic extraction into a hierarchical tree structure, where the entire image is treated as the root node and key objects within the image serve as the first-level child nodes under the root. Each object node is then connected to a set of attribute leaf nodes, such as color, shape, material, and spatial position. Through this hierarchical tree modeling, object-level semantics and finer-grained attribute-level semantics are progressively integrated into the image description. At the receiver, we further adapt the LLM-grounded Diffusion (LMD) method \cite{lmd} so that it can directly use the structured text description provided by TSASE.

Second, as illustrated in Fig.~\ref{fig:comparison}b, we introduce Initial Noise Optimization (INO) to select an initial noise that is more favorable for semantically faithful reconstruction. The transmitter is equipped with an image generation model identical to the one at the receiver side. Before transmitting an image's text description, the transmitter tests multiple candidate noise seeds and selects the one that yields the highest reconstruction similarity. This selected noise seed, together with the text description, is delivered to the receiver for image reconstruction.

\begin{figure*}[t]
    \centering
    \includegraphics[width=0.85\textwidth]{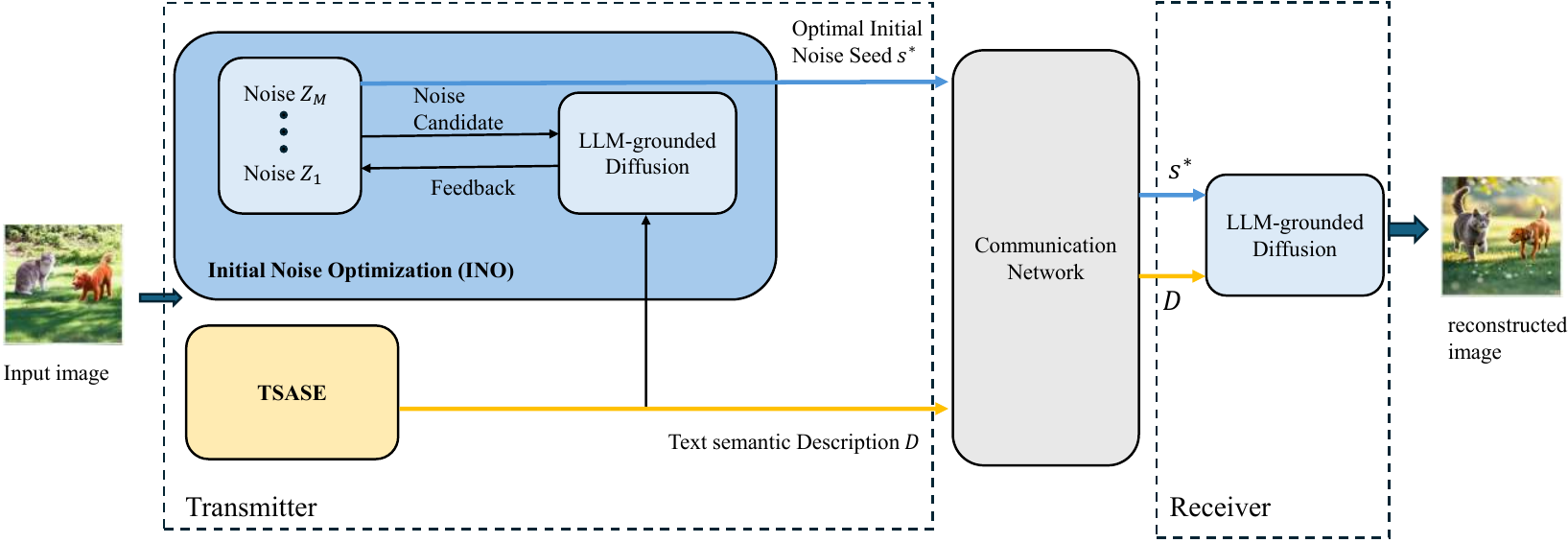}
    \caption{Overall architecture of the proposed TISC framework. At the transmitter, TSASE extracts the structured semantic description $D$, while INO selects the initial noise seed $s^*$. These two components are introduced in Sections~\ref{sec:q1} and~\ref{sec:q2}, respectively. The communication network delivers $D$ and $s^*$ to the receiver, where they are used to guide LLM-grounded Diffusion-based image reconstruction.}
    \label{fig:architecture}
\end{figure*}

\textbf{Related work:} Existing image semantic communication approaches can be grouped into two lines of research. The first delivers compressed image features, such as semantic segmentation maps \cite{OOSC,semanticmap2,semanticmap3}, latent representations \cite{latent1,Deepjscc,NTSCC,saoosc,transformerjscc,swinjscc,deepjscc-f}, or hierarchical semantic features \cite{semanticmap4}, between the transmitter and the receiver. Although these approaches significantly reduce bandwidth consumption compared to conventional pixel-wise image transmission, the compressed image features may still incur non-negligible communication overhead.

This paper belongs to the second line of research, which delivers the text modality between the transmitter and the receiver for highly efficient bandwidth utilization. In this image-text-image paradigm, the semantic content of an image is extracted as text information, and the receiver reconstructs the image using generative models. Existing studies in this direction typically employ MLLMs or image captioning models to directly generate holistic text descriptions from input images \cite{txt_sem1,txt_sem2,txt_sem3}. For these studies, the semantic extraction performance mainly depends on the inherent capability of the adopted model; there is no explicit optimization of the semantic extraction process for faithful image description. As a result, semantic distortions can be introduced due to potential inaccuracies in the text descriptions. In addition, the lack of explicit spatial information may result in the receiver placing objects at wrong positions. This paper aims to address these problems.

Following the second line of research, we emphasize that this paper focuses on compressing an image source into a compact and semantically faithful text representation for transmission, and on reconstructing the image from this representation at the receiver using a generative model. The problem studied in this paper is therefore more closely related to the source coding and decoding process in communication systems: the transmitter determines which reconstruction-relevant information should be preserved and encoded while removing redundant content, and the receiver recovers the target image based on the transmitted information. Accordingly, we position TISC as a text-driven semantic source coding and decoding framework. Under this positioning, we assume that the transmitted text information can be reliably delivered through the underlying communication network, for example, via conventional channel coding, retransmission mechanisms, or TCP-like reliable transport protocols \cite{tcp}.

%% file: sections/02_tsase.tex
\section{I2T: Addressing Question 1}
\label{sec:q1}

To address Question 1, we propose TSASE, which extracts a structured and semantically faithful text description from the original image to reduce information loss and distortion during I2T semantic extraction. As shown in the overall TISC framework in Fig.~\ref{fig:architecture}, this extraction process is performed at the transmitter, where the input image is converted into a text-based semantic description. This description is then used as the semantic condition for receiver-side image reconstruction. This section focuses on the design of TSASE. Subsection~\ref{subsec:tsase} presents the TSASE process and defines the organization of its structured semantic output. Subsection~\ref{subsec:lmd_adapt} explains how the receiver uses the TSASE description for image reconstruction.

\subsection{Tree-Structured Attribute Semantic Extraction}
\label{subsec:tsase}

Straightforward approaches \cite{txt_sem1,txt_sem2,txt_sem3} obtain an image description by directly feeding the entire image to MLLMs. When dealing with complex inputs, such as images with multiple objects, these methods often overlook subtle semantic details, leading to inaccurate image descriptions at the transmitter side. TISC addresses this issue with the TSASE paradigm introduced as follows.

TSASE decomposes a complex image description task into a set of simpler subtasks, with each subtask aiming to clearly describe one object within the image. For each object-specific subtask, its description is further decomposed into several attribute-specific mini-tasks, with each of these mini-tasks focusing on one aspect of the object’s attribute (e.g., spatial position, shape, color, material, etc.). With the completion of all these attribute-specific mini-tasks, we obtain detailed descriptions for an object. In addition to these detailed object descriptions, a global scene of the image and a specific description focusing on the background are also included in the image description to ensure faithful reconstruction at the receiver side.

\begin{figure*}[t]
    \centering
    \includegraphics[width=0.90\textwidth]{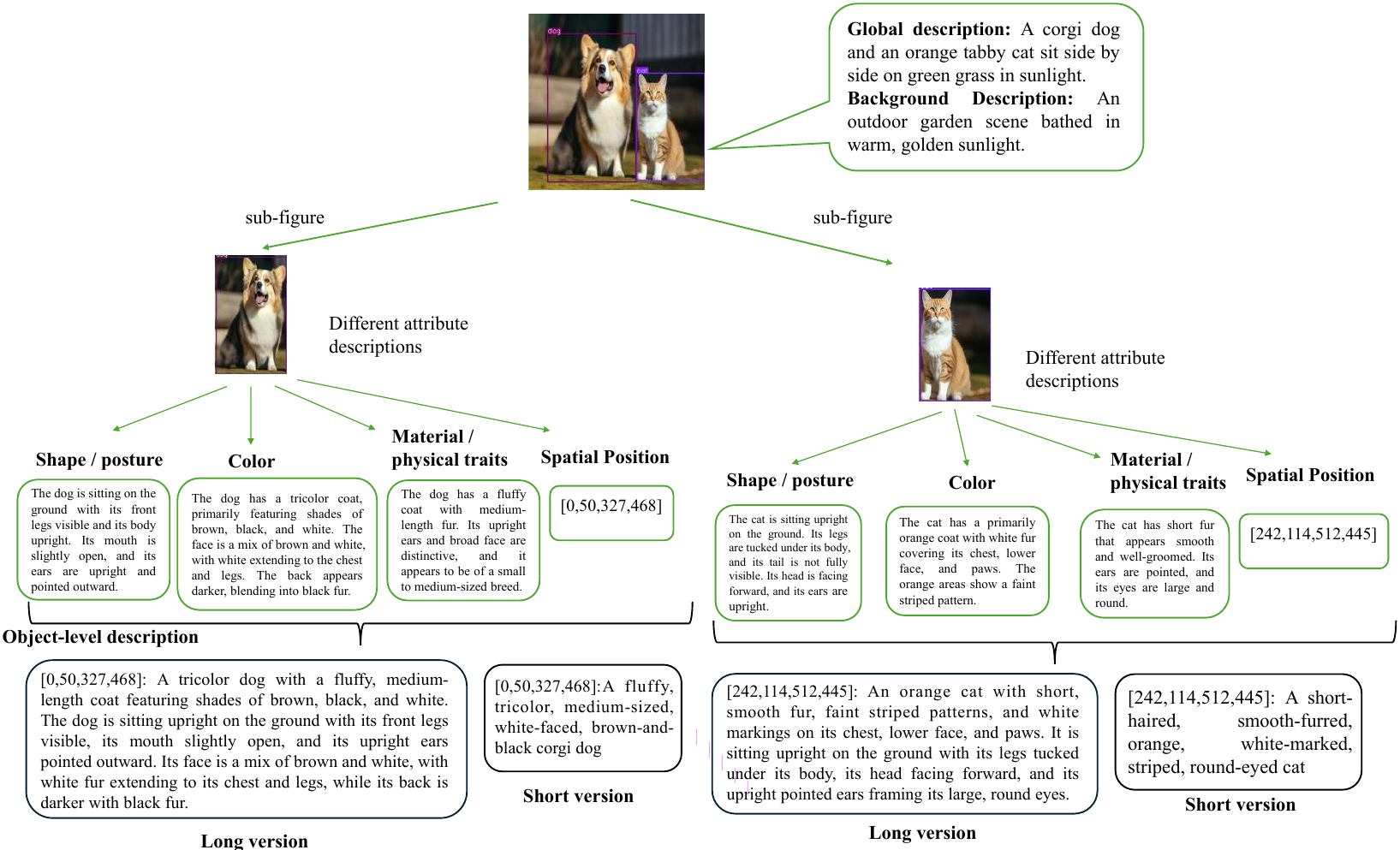}
    \caption{Example of the tree-structured attribute semantic extraction process.}
    \label{fig:tsase_tree}
\end{figure*}

We now elaborate the TSASE paradigm with Fig.~\ref{fig:tsase_tree}. At the beginning of the semantic extraction process, we take the whole image as the ``root node'' of the tree structure. We give the whole image to an MLLM and ask the model to generate the root node, which includes the image's global description $D_{\mathrm{global}}$ and the image's background description $D_{\mathrm{bg}}$. In Fig.~\ref{fig:tsase_tree}, for example, $D_{\mathrm{global}}$ highlights the primary content and relationships in the image with description ``\textit{A corgi dog and an orange tabby cat sit side by side on green grass in sunlight}'', while $D_{\mathrm{bg}}$ emphasizes the background of the image with description ``\textit{An outdoor garden scene bathed in warm, golden sunlight}''.

After obtaining the global scene information, TSASE employs YOLOv11~\cite{yolov11x} to detect and localize all objects within the image. Let us assume $N$ objects within the given image. The set of rectangular bounding boxes for detected objects are denoted by $\bm{B}=\langle \bm{b_1},\ldots,\bm{b_N}\rangle$, where bounding box $\bm{b_i}=(x_i^{\mathrm{tl}},y_i^{\mathrm{tl}},x_i^{\mathrm{br}},y_i^{\mathrm{br}})$ specifies the top-left and bottom-right coordinates of object $i$. The origin $(0,0)$ of the image coordinate system is defined at the top-left corner of the whole image. To prevent mistaken object detections, such as multiple bounding boxes for a single object, we have a pre-defined confidence threshold for the object detection, which is used to compare with the confidence score that YOLOv11 provided for each detected object, to filter out low-confidence detections that are likely to be problematic.

Based on the remaining bounding boxes, the whole image is cropped into $N$ single-object sub-images, denoted by $\{I_i^{\mathrm{obj}}\}_{i=1}^{N}$, which are then used to extract the fine-grained visual attributes of the corresponding objects. These sub-images serve as first-level child nodes of the original image. In TSASE, we have a pre-defined list of finer-grained attributes of interest to us, which include the object's spatial position, shape/pose, color, and material/physical attributes\footnote{The specific division of attributes can be adjusted according to the focus of the downstream task. For example, if the downstream application places greater emphasis on object color and shape, the attribute dimensions can be correspondingly expanded or refined. The attribute categorization presented here serves merely as a proof of concept and can be further customized based on practical requirements.}. For sub-image $I_i^{\mathrm{obj}}$ corresponding to the $i$-th first-level child node, these attributes of that sub-image serve as the second-level leaf nodes. Among these attributes, the spatial-position leaf node is directly given by the bounding box $\bm{b_i}$ obtained from YOLOv11. For example, the spatial attribute of the first object (i.e., dog in Fig.~\ref{fig:tsase_tree}) is $\bm{b_1}=(0,50,327,468)$. For each remaining attribute leaf node, we ask the MLLM to describe one attribute of sub-image $I_i^{\mathrm{obj}}$ at one time, helping the MLLM to concentrate on one specific aspect of the sub-image in one interaction. In Fig.~\ref{fig:tsase_tree}, the shape/pose description of the first sub-image could be: ``The dog is sitting on the ground with its front legs visible and its body upright. Its mouth is slightly open, and its ears are upright and pointed outward.''

\begin{figure}[t]
    \centering
    \includegraphics[width=0.5\textwidth]{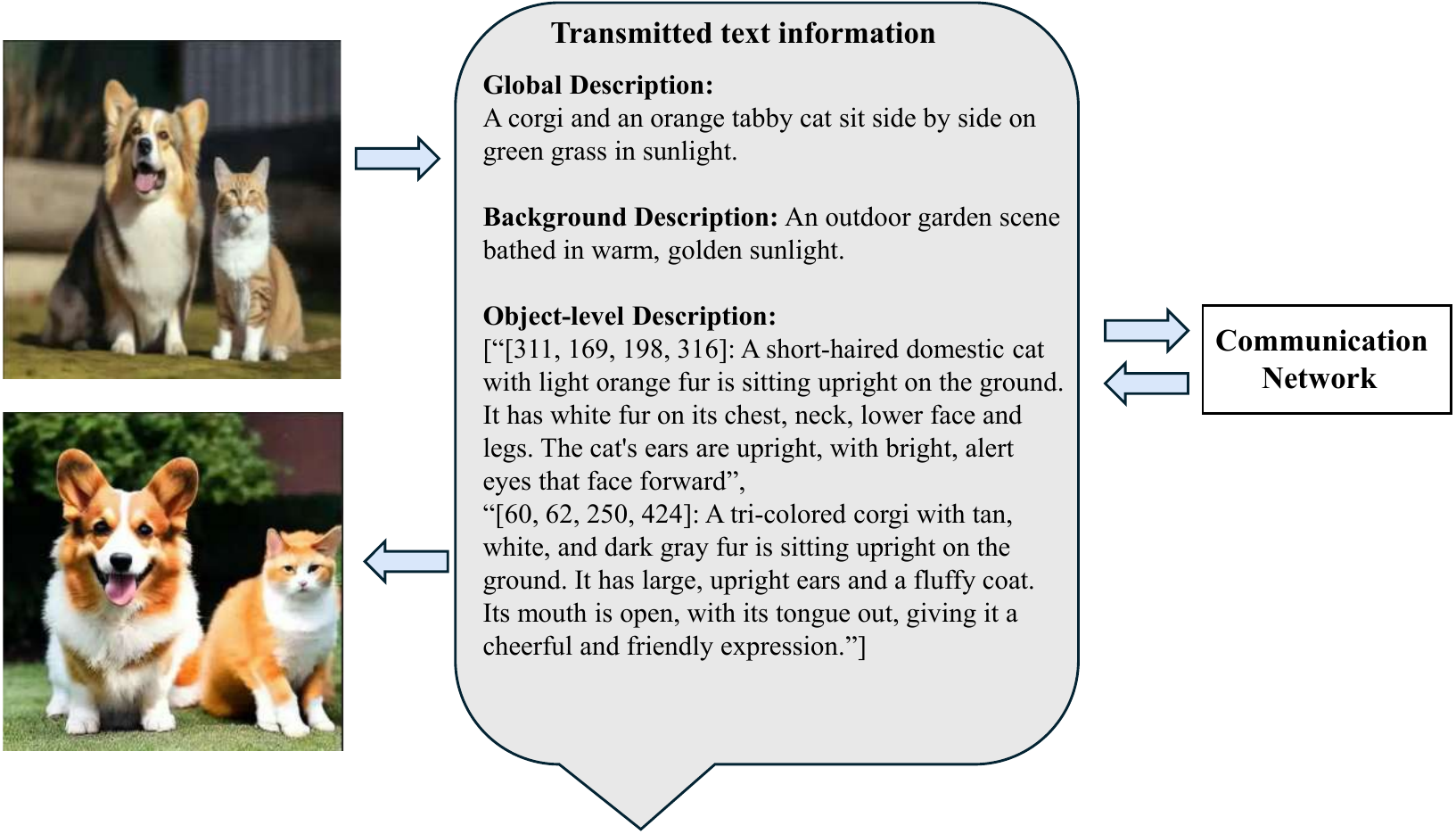}
    \caption{An example of the structured semantic description produced by TSASE.}
    \label{fig:tsase_output}
\end{figure}

After all attribute nodes are obtained, TSASE integrates the spatial-position attribute and the MLLM-generated attribute descriptions of the $i$-th object to form a semantic representation $d_i$. Note that $d_i$ is a simple concatenation of attribute descriptions, which can be rather lengthy (see the ``long version'' object description illustrated in Fig.~\ref{fig:tsase_output}). To reduce information redundancy, we leverage an LLM (e.g., GPT-4o \cite{gpt4o} in this paper) to further refine the long version into a ``short version'' object description. For example, for the first sub-image in Fig.~\ref{fig:tsase_tree}, its long version object description ``[0,50,327,468]: \textit{A tricolor dog with a fluffy, medium-length coat featuring shades of brown, black, and white. The dog is sitting upright on the ground with its front legs visible\ldots}'' can be compressed as ``[0,50,327,468]: \textit{A fluffy, tricolor, medium-sized, white-faced, brown-and-black corgi dog}''. In our experiments, we find that short version descriptions can achieve generation results no worse than longer ones. Hence, this paper uses the short description of object $i$, which is denoted as $d_i^S$, as the object semantic representation.

With the short-version description for each object, we obtain the object-level representation of the whole image as $\bm{D_{\text{obj}}}=[d_1^S,\ldots,d_N^S]$. Finally, $D_{\text{global}}$, $D_{\mathrm{bg}}$, and $\bm{D_{\text{obj}}}$ are concatenated to form a comprehensive image description $\bm{D}=\{D_{\text{global}},D_{\mathrm{bg}},\bm{D_{\text{obj}}}\}$.

To further clarify the practical organization of the TSASE output, Fig.~\ref{fig:tsase_output} provides an example of the structured semantic description $\bm{D}$. As illustrated in Fig.~\ref{fig:tsase_output}, the information is organized as follows:
\begin{enumerate}
    \item the beginning of the image description includes $D_{\text{global}}$, which provides an overall summary of the image content.
    \item the background description $D_{\mathrm{bg}}$ supplements contextual information of the environment, facilitating more accurate reconstruction of the image background.
    \item object-level descriptions $\bm{D_{\text{obj}}}$ are presented together.
\end{enumerate}

After obtaining $\bm{D}$, the transmitter treats it as the text-based semantic information to be delivered. In this paper, we assume that the transmitted text information can be reliably delivered through the communication network, e.g., via TCP. As shown by the yellow arrow in Fig.~\ref{fig:architecture}, $\bm{D}$ is transmitted to the receiver and used as the semantic condition for image reconstruction.

\subsection{TSASE-Guided Image Reconstruction}
\label{subsec:lmd_adapt}

As shown in Fig.~\ref{fig:architecture}, the receiver obtains the TSASE output $D$ through the reliable communication network. The receiver-side image reconstruction in TISC is built upon the open-source LLM-grounded Diffusion (LMD) model~\cite{lmd}, and is jointly guided by $\bm{D}$ and the noise seed $s^*$. This subsection focuses on how $\bm{D}$ is adapted to the input of LMD, such that the global scene description, background description, and object-level attribute descriptions can be used for image reconstruction. The selection of $s^*$ and its role in providing the initial noise for the LMD generation process will be discussed in Section~\ref{sec:q2}. We only briefly introduce the LMD functions relevant to this adaptation, without delving into its implementation details or underlying mechanisms (we refer interested readers to Section~III in~\cite{lmd} for details).

To clarify how LMD is adapted in our framework, we first briefly review the original LMD generation pipeline. In generation, the image reconstruction pipeline in original LMD can be divided into two major steps: 1) LLM-driven reasoning of semantic information, and 2) diffusion-based image generation with object-level semantic and spatial alignment.

For the first step, LMD employs an LLM to interpret the text description input. The LLM parses object information, background information, and relative spatial relationships among objects from the free-form text. Leveraging its contextual understanding and prior knowledge of the real world, the LLM estimates the spatial coordinates of different objects and generates the description of each object, the background description of the image, and the global scene description of the image to enrich them accordingly.

For the second step, the enriched semantic information, along with a randomly generated noise vector, is given to a diffusion model for guiding the image generation. During this process, LMD first generates the semantic representation of each object based on its description. To maintain spatial alignment, each object's semantic representation will be placed at a location predefined in its spatial description. The above object-level treatments ensure semantic and spatial alignment for all objects within the image. Following object-level treatments, LMD generates the whole image based on the representation of each object, the background and the global descriptions of the image, and the initial noise vector.

We further clarify the difference between our adaptation of the LMD input flow and the original LMD framework. In the original LMD, step~(1) is performed by the LLM to generate semantic information from the input text description via model reasoning. In our framework, however, $\bm{D}$ already contains precise semantic details extracted from the original image, making the LLM reasoning in step~1 unnecessary---we directly use $\bm{D}$ as the input for step~2 of the original LMD pipeline.

%% file: sections/03_ino.tex
\section{T2I: Addressing Question 2}
\label{sec:q2}

As discussed in Section~\ref{sec:introduction}, even when the receiver obtains a semantically faithful text description, diffusion-based reconstruction may still produce images with different levels of semantic faithfulness due to different initial noises. Therefore, Question~2 focuses on how to select an initial noise that is more favorable for reconstructing the semantic content of the original image.

To address this question, we introduce an Initial Noise Optimization (INO) mechanism at the transmitter. INO tests multiple candidate initial noises before transmission and uses the comprehensive similarity score $S_{\mathrm{comp}}$ to evaluate the consistency between each corresponding reconstruction and the original image. Here, $S_{\mathrm{comp}}$ is an image similarity metric defined in this paper, which jointly considers visual similarity and semantic similarity. INO then selects the initial noise with the highest $S_{\mathrm{comp}}$ score to support more semantically faithful image reconstruction at the receiver.

This section introduces INO from two aspects. Section~\ref{subsec:ino} first presents the basic pipeline of INO. Section~\ref{subsec:scomp} then defines the proposed similarity evaluation metric $S_{\mathrm{comp}}$ used in INO.

\subsection{Initial Noise Optimization}
\label{subsec:ino}

At the transmitter side, we deploy an image generation model $G(\cdot)$ that is identical to the one at the receiver. The aim of doing so is to assess the expected image reconstruction performance under different initial noises before the transmission process. Let us assume that the transmitter tests $m$ noise vectors, denoted by $Z=[z_1,\ldots,z_m]$, to select the best-performing initial noise among these candidates. The choice of $m$ controls the search budget of INO and will be determined through the parameter study in Experiment~1 of Section~\ref{subsec:InoParam}. For each $z_q \in Z$, conditioned on the semantic description $\bm{D}$ produced by TSASE, the reconstructed image is generated as $\hat{I}_q=G(D,z_q)$.

At the transmitter, we evaluate the similarity between the original image $I$ and each candidate reconstruction $\hat{I}_q$ using the proposed comprehensive similarity score $S_{\mathrm{comp}}(\cdot,\cdot)$ (see Section~\ref{subsec:scomp} and Eq.~\ref{eq:scomp} for more details). Among all possible noise vectors, the one resulting in the highest similarity with $I$ is selected, i.e.,
\begin{equation}
z^*=\arg\max_{z_q \in Z} S_{\mathrm{comp}}(I,\hat{I}_q).
\label{eq:zstar}
\end{equation}

After finding the best-performing noise vector $z^*$, the next issue is how to convey this selected noise to the receiver efficiently. We note that there is no need to transmit the complete noise vector for communication efficiency---transmitting the random seed for generating the noise is enough. Let us denote the random noise seed that one-to-one maps to $z^*$ by $s^*$ and write it as $s^*=f(z^*)$, where $f$ denotes the mapping function.

Subsequently, as indicated by the blue arrow in Fig.~\ref{fig:architecture}, the selected seed $s^*$ is transmitted to the receiver through the communication network. Here, $s^*$ is essentially a random seed used to generate the initial noise $z^*$, and thus can be transmitted as text-form information with very low communication overhead. In our receiver-side pipeline, we set $z^*$ as the starting point of the LMD diffusion generation process, and use the semantic description $\bm{D}$ as the structured semantic input to LMD for image reconstruction, following the method described in Section~\ref{subsec:lmd_adapt}.

\subsection{Comprehensive Similarity Metric for Candidate Noise Evaluation}
\label{subsec:scomp}

In INO, the comprehensive similarity score $S_{\mathrm{comp}}(I,\hat{I}_q)$ is used to measure the consistency between the original image $I$ and the $q$-th candidate reconstruction $\hat{I}_q$. This score indicates whether the corresponding candidate initial noise $z_q$ is favorable for reconstructing the semantic content of the original image. Since TISC aims at semantically faithful image reconstruction, this consistency should be evaluated from both visual and semantic perspectives. Therefore, we define two components, namely the visual similarity $S_{\mathrm{vis}}$ and the semantic similarity $S_{\mathrm{sem}}$, and then construct $S_{\mathrm{comp}}$ based on them.

To measure visual consistency, we adopt the Learned Perceptual Image Patch Similarity (LPIPS) distance~\cite{lpips}. LPIPS quantifies the perceptual distance between two images using a pretrained convolutional neural network (CNN), such as AlexNet~\cite{alexnet} or VGG~\cite{VGG}. Let us denote the network by $F(\cdot)$, and we write the network's output embedding at layer $l$ as $F_l(I)$. Given $I$ and $\hat{I}_q$, the LPIPS of the two images is given by:
\begin{equation}
\mathrm{LPIPS}(I,\hat{I}_q)=\sum_l w_l \cdot \left\|F_l(I)-F_l(\hat{I}_q)\right\|_2^2.
\label{eq:lpips}
\end{equation}

where $w_l$ is a predefined weight of layer $l$ within the network, and $\left\|\cdot\right\|_2^2$ denotes the normalized Euclidean distance between two vectors. A lower $\mathrm{LPIPS}(I,\hat{I}_q)$ indicates that the two images are closer in the perceptual feature space. To make the visual term consistent with the subsequent semantic similarity term, where a larger value indicates higher similarity, we convert the LPIPS distance into a visual similarity score:
\begin{equation}
S_{\mathrm{vis}}(I,\hat{I}_q)=1-\mathrm{LPIPS}(I,\hat{I}_q).
\label{eq:svis}
\end{equation}
Therefore, a larger $S_{\mathrm{vis}}$ indicates higher perceptual similarity between the original image and the candidate reconstruction.

For semantic similarity, we focus on the similarity between the tree-structured semantic descriptions of the original image $I$ and the $q$-th candidate reconstruction $\hat{I}_q$. To this end, we use TSASE to extract structured descriptions from both images, and further compare their text similarities at three levels: the global scene, the background, and object-level semantics. Specifically, for the original image $I$, TSASE extracts the global scene description $D_{\mathrm{global}}$, the background description $D_{\mathrm{bg}}$, and object-level semantic descriptions $\bm{D_{\mathrm{obj}}}=[d_1^S,\ldots,d_{n_1}^S]$, where $d_i^S$ denotes the short-version structured description of the $i$-th detected object (see Section~\ref{subsec:tsase} for more details). For the corresponding reconstructed image $\hat{I}_q$, TSASE similarly extracts the corresponding descriptions $D_{\mathrm{global}}'$, $D_{\mathrm{bg}}'$, and $\bm{D_{\mathrm{obj}}}'=[{d'}_1^S,\ldots,{d'}_{n_2}^S]$, where $n_1$ and $n_2$ denote the numbers of detected objects in $I$ and $\hat{I}_q$, respectively.

To measure the semantic closeness between these text descriptions, we adopt Sentence-BERT (SBERT)~\cite{sbert} as the text similarity model. SBERT encodes each text description into a semantic embedding vector, and the similarity between two descriptions is computed based on the cosine similarity of their embeddings. In this paper, $\mathrm{SBERT}(\cdot,\cdot)$ denotes the normalized SBERT-based similarity score in $[0,1]$, where a larger value indicates higher semantic similarity.

Based on this definition, the global scene similarity and the background similarity are computed as $\mathrm{SBERT}(D_{\mathrm{global}},D_{\mathrm{global}}')$ and $\mathrm{SBERT}(D_{\mathrm{bg}},D_{\mathrm{bg}}')$, respectively. For object-level semantic similarity, it is not appropriate to directly compare the object-level descriptions according to the detection order of the corresponding objects, because the detected object sets in the original image and the candidate reconstruction may differ in both size and order. Therefore, we first establish object correspondences between the two images according to their spatial locations. These correspondences determine which object-level descriptions can be compared in a one-to-one manner.

\begin{figure}[t]
    \centering
    \includegraphics[width=0.92\columnwidth]{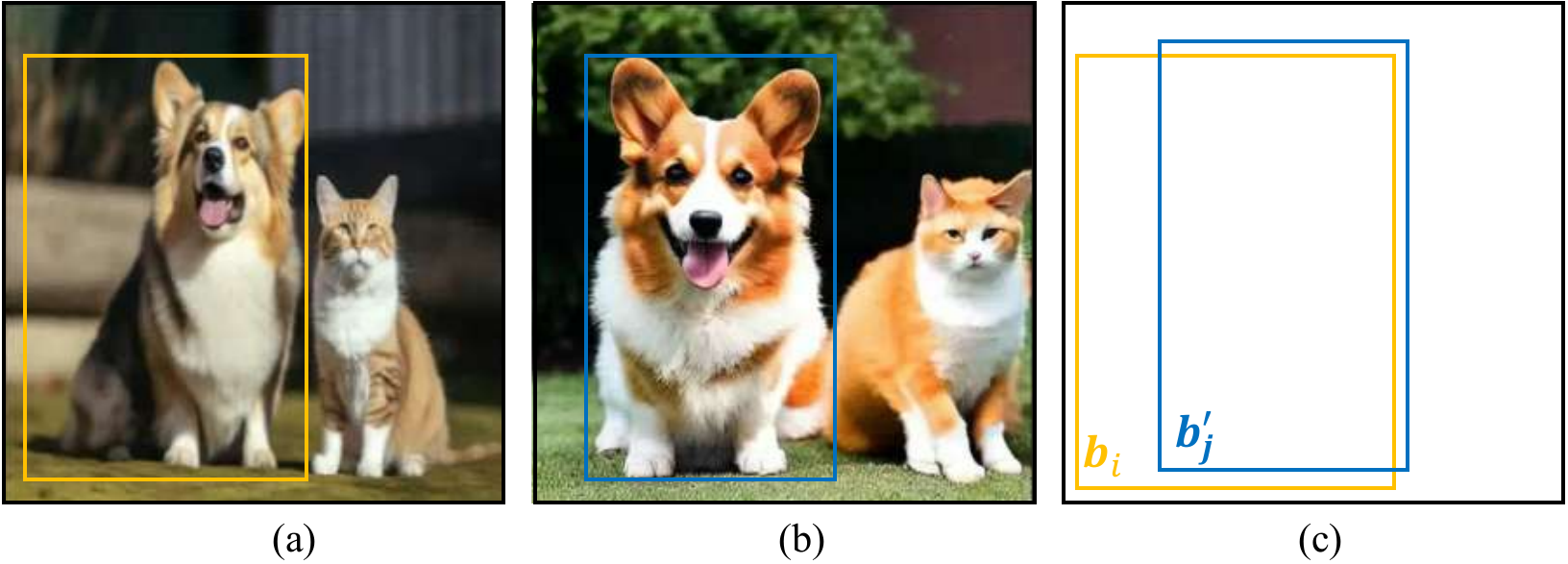}
    \caption{Illustration of IoU calculation for spatial matching. (a) Original image, where the yellow box denotes the bounding box $\bm{b_i}$ provided by the spatial-position attribute of the $i$-th object description $d_i^S$. (b) Candidate reconstruction, where the blue box denotes the bounding box $\bm{b_j'}$ provided by the spatial-position attribute of the $j$-th object description ${d'}_j^S$. (c) Spatial overlap between $\bm{b_i}$ and $\bm{b_j'}$, which is used to compute their IoU.}
    \label{fig:iou_calculation}
\end{figure}

To establish this spatial matching relationship, we compute the Intersection over Union (IoU) using the bounding-box coordinates contained in the object-level descriptions. IoU is a commonly used metric for measuring the spatial overlap between two bounding boxes~\cite{iou}. Specifically, for the $i$-th object description $d_i^S$ from the original image, its spatial-position attribute provides the bounding box $\bm{b_i}$; for the $j$-th object description from the candidate reconstruction, its spatial-position attribute provides the bounding box $\bm{b_j'}$. Fig.~\ref{fig:iou_calculation} presents an illustrative example of the IoU calculation, where the IoU between $\bm{b_i}$ and $\bm{b_j'}$ is computed as:
\begin{equation}
\mathrm{IoU}=\frac{\mathrm{area}(\bm{b_i}\cap \bm{b_j'})}{\mathrm{area}(\bm{b_i}\cup \bm{b_j'})}.
\label{eq:iou}
\end{equation}
Here, $(\bm{b_i}\cap \bm{b_j'})$ is the overlap area and $(\bm{b_i}\cup \bm{b_j'})$ is the union area. A larger IoU indicates a higher degree of spatial overlap between the two detected objects. Following the commonly used criterion in object detection evaluation~\cite{iou}, we perform spatial matching between detected objects in the original image and the candidate reconstruction. Two detected objects are regarded as spatially matched if their IoU is larger than $0.5$.

\begin{figure}[t]
    \centering
    \includegraphics[width=0.475\textwidth]{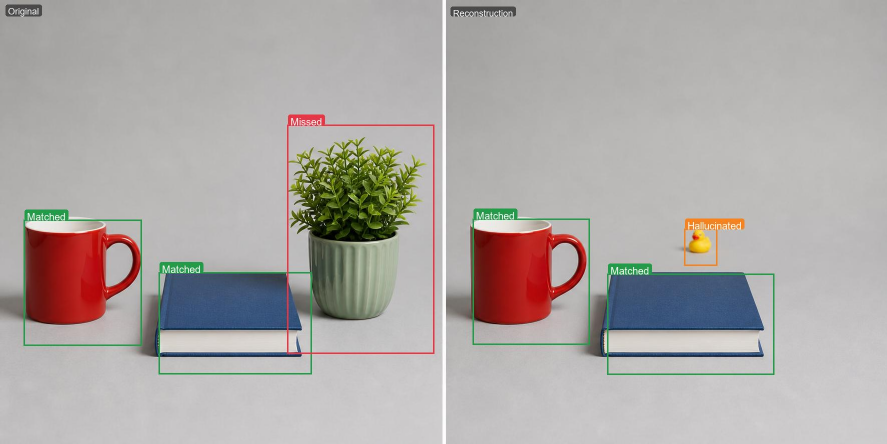}
    \caption{Illustration of matched, missed, and hallucinated objects based on spatial matching. The left panel shows the original image, and the right panel shows the reconstructed image. Green boxes indicate matched objects that appear in both images with sufficient spatial overlap. The red box indicates a missed object that appears only in the original image, while the orange box indicates a hallucinated object that appears only in the reconstructed image.}
    \label{fig:matched_missed_hallucinated}
\end{figure}

Based on this spatial matching result, the objects in the original and reconstructed images are divided into three categories: matched, missed, and hallucinated objects. Fig.~\ref{fig:matched_missed_hallucinated} shows an illustrative example. Matched objects refer to object pairs that appear in both images and are spatially matched (i.e., the red mug and the blue book in Fig.~\ref{fig:matched_missed_hallucinated}). We use $N_{\mathrm{mat}}$ to denote the number of matched object pairs (e.g., $N_{\mathrm{mat}}=2$ in Fig.~\ref{fig:matched_missed_hallucinated}), and $M=\{i_k,j_k\}_{k=1}^{N_{\mathrm{mat}}}$ to denote the set of all matched object pairs, where $i_k$ and $j_k$ index the $k$-th matched objects in the original and reconstructed images, respectively. For the $k$-th matched object pair $(i_k,j_k)$ in $M$, we compute its object-level semantic similarity as $\mathrm{SBERT}(d_{i_k}^S,{d'}_{j_k}^S)$. Missed objects refer to objects that appear in the original image but have no spatially matched counterpart in the reconstructed image (i.e., the plant in Fig.~\ref{fig:matched_missed_hallucinated}), and their number is denoted by $N_{\mathrm{miss}}$ (e.g., $N_{\mathrm{miss}}=1$ in Fig.~\ref{fig:matched_missed_hallucinated}), and hallucinated objects refer to extra objects generated in the reconstructed image but absent from the original image (i.e., the toy duck in Fig.~\ref{fig:matched_missed_hallucinated}), and their number is denoted by $N_{\mathrm{hal}}$ (e.g., $N_{\mathrm{hal}}=1$ in Fig.~\ref{fig:matched_missed_hallucinated}). The similarity scores of both missed and hallucinated objects are set to zero.

Based on the above definitions, the final TSASE-based semantic similarity is computed as the average similarity over the global scene, the background, and all object-level semantic descriptions:
\begin{equation}
\begin{aligned}
S_{\mathrm{sem}}(I,\hat{I}_q)
=
\frac{
\begin{aligned}
&\mathrm{SBERT}(D_{\mathrm{global}},D'_{\mathrm{global}})
+\mathrm{SBERT}(D_{\mathrm{bg}},D'_{\mathrm{bg}}) \\
&+\sum_{k=1}^{N_{\mathrm{mat}}}
\mathrm{SBERT}(d_{i_k}^{S},{d'}_{j_k}^{S})
\end{aligned}
}{
N_{\mathrm{mat}}+N_{\mathrm{miss}}+N_{\mathrm{hal}}+2
}.
\end{aligned}
\label{eq:ssem}
\end{equation}

Here, the additional term $+2$ in the denominator corresponds to the global scene description and the background description, while $N_{\mathrm{mat}}$, $N_{\mathrm{miss}}$, and $N_{\mathrm{hal}}$ correspond to the numbers of matched, missed, and hallucinated objects, respectively. Since missed and hallucinated objects are assigned zero similarity scores, they do not appear in the numerator but are included in the denominator, thereby reducing the semantic similarity score. Compared with a single global-caption similarity metric, this TSASE-based metric evaluates semantic faithfulness from three levels: the overall scene, the background information, and object-level semantic details. It also explicitly reflects the impact of missing objects and hallucinated objects in the reconstructed image.

After obtaining the visual similarity $S_{\mathrm{vis}}(I,\hat{I}_q)$ and the semantic similarity $S_{\mathrm{sem}}(I,\hat{I}_q)$, we fuse them with a weighted sum to obtain the comprehensive similarity score for candidate noise evaluation:
\begin{equation}
S_{\mathrm{comp}}(I,\hat{I}_q)=\alpha S_{\mathrm{vis}}(I,\hat{I}_q)+(1-\alpha)S_{\mathrm{sem}}(I,\hat{I}_q).
\label{eq:scomp}
\end{equation}
Here, $\alpha\in[0,1]$ controls the relative weights of the visual and semantic similarities, and its value is determined through the parameter study (see Experiment~2 in Section~\ref{subsec:InoParam} for more details). Since both $S_{\mathrm{vis}}(I,\hat{I}_q)$ and $S_{\mathrm{sem}}(I,\hat{I}_q)$ lie in $[0,1]$, $S_{\mathrm{comp}}(I,\hat{I}_q)$ also lies in $[0,1]$. A larger $S_{\mathrm{comp}}$ indicates higher consistency between the candidate reconstruction and the original image in both visual and semantic aspects.

%% file: sections/04_experiments.tex
\section{Experimental Results}
\label{sec:experiments}

This section evaluates the key designs of the proposed TISC framework through experiments. The experiments are organized into two subsections. Section~\ref{subsec:tsase_ablation} conducts an ablation study on the TSASE module to evaluate its effectiveness in semantic extraction. Specifically, we assess the contribution of TSASE from two perspectives: spatial-attribute recovery and the semantic faithfulness of the extracted descriptions. Section~\ref{subsec:InoParam} further analyzes the parameter settings of the INO mechanism, focusing on the number of candidate initial noises $m$ and the weighting factor $\alpha$ between visual similarity and semantic similarity in the comprehensive similarity score $S_{\mathrm{comp}}$. The datasets used in these experiments are summarized as follows.

\textbf{Tested Datasets:} we tested five representative image datasets with complementary characteristics, covering diverse visual scenes and object complexities. Tested dataset includes COCO \cite{coco}, Visual Genome V1.2 \cite{VG}, Flickr30K \cite{flickr}, Open Images V7 \cite{openimage}, and the ILSVRC2012 validation set \cite{ILSVRC}.

\subsection{TSASE Ablation Study}
\label{subsec:tsase_ablation}

This subsection evaluates the semantic extraction capability of the TSASE module. To this end, we compare the complete TISC framework with the following ablation setting:

\textbf{w/o TSASE:} The TSASE module is removed. In this setting, the transmitter no longer extracts tree-structured descriptions. Instead, an MLLM is directly used to generate a one-shot holistic description of the entire image. To focus the comparison on the role of TSASE, all other system settings remain unchanged. The receiver still uses the noise seed selected by INO and reconstructs the image through the LMD model based on this alternative text description. Note that this setting is consistent with the holistic-description-based image-text-image pipeline adopted by the related works~\cite{txt_sem1,txt_sem2,txt_sem3} discussed in Section~\ref{sec:introduction}. Therefore, while this setting is designed as an ablation case for isolating the role of TSASE, it can also be viewed as a representative benchmark setting for these prior text-driven semantic communication systems.

\begin{figure}[t]
    \centering
    \includegraphics[width=\linewidth]{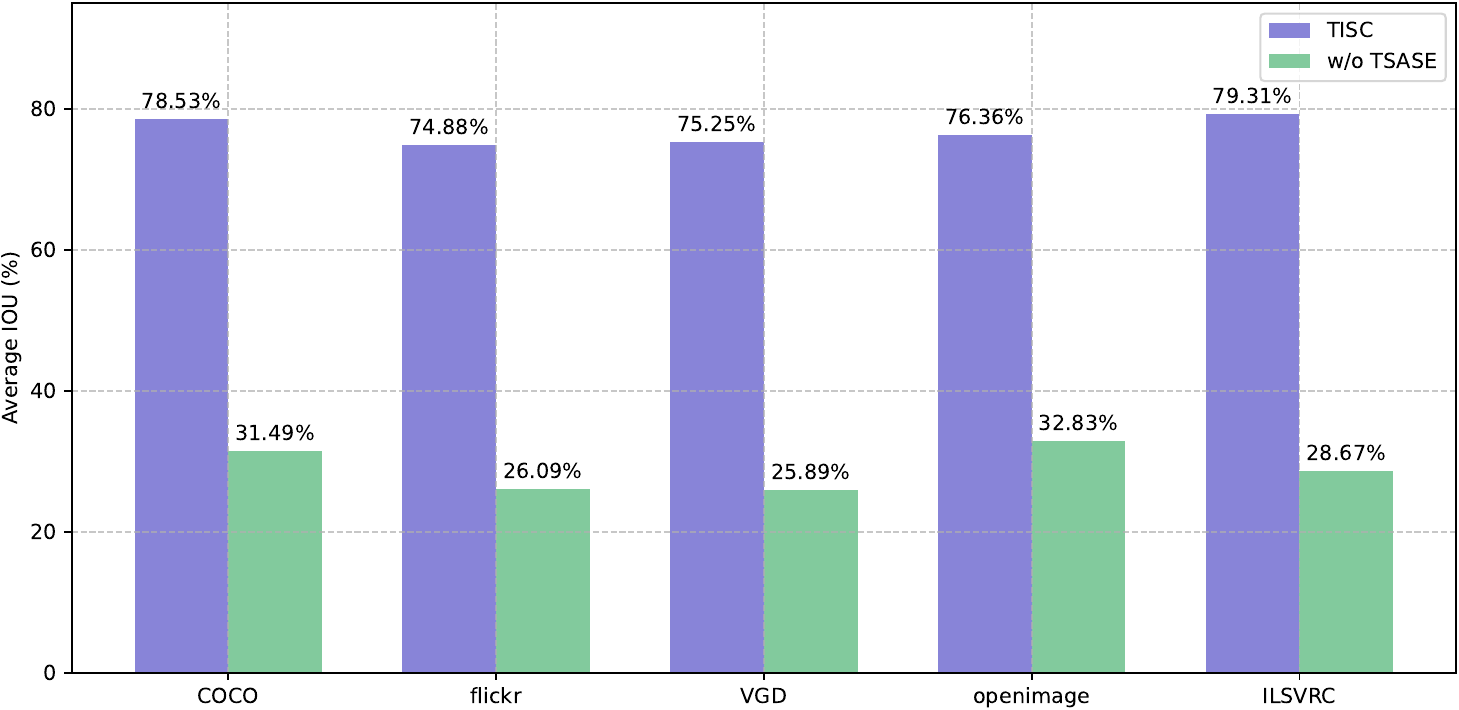}
    \caption{Comparison of average IoU between TISC and the w/o TSASE ablation across tested datasets.}
    \label{fig:avg_iou}
\end{figure}

TSASE extracts a structured semantic description from the original image, which includes not only fine-grained visual attributes such as color, shape, and material, but also the spatial-position attributes of objects. Therefore, we evaluate its semantic extraction capability from two perspectives. First, we examine whether the spatial attributes extracted by TSASE help improve the recovery of object positions in the reconstructed image. Second, we evaluate whether TSASE can more accurately extract semantic information beyond spatial position.

We first evaluate the recovery of spatial attributes. Given an original image transmitted and the image recovered at the receiver side under the ablation setup, we evaluate the spatial accuracy of the reconstructed image with IoU (see Eq.~\eqref{eq:iou} for more details).

For each test dataset, we compute the average IoU over all matchable objects to measure the overall recovery of object positions in the reconstructed images. Let us assume a total of $n$ objects in the dataset for calculating the IoU, then the average IoU of a tested dataset can be expressed as
\begin{equation}
\overline{\mathrm{IoU}}=\frac{1}{n}\sum_{i=1}^{n}\mathrm{IoU}_i,
\label{eq:avg_iou}
\end{equation}
where $\mathrm{IoU}_i$ denotes the IoU of the $i$-th detected object.

Fig.~\ref{fig:avg_iou} benchmarks the average IoU of TISC and the w/o TSASE ablation across five tested datasets. The figure shows that TISC consistently achieved higher average IoUs than w/o TSASE across all datasets---with most datasets exceeding $75\%$ (e.g., $78.53\%$ on COCO, $79.31\%$ on ILSVRC), while the ablation case is around $30\%$ in general.

\begin{figure}[t]
    \centering
    \includegraphics[width=\linewidth]{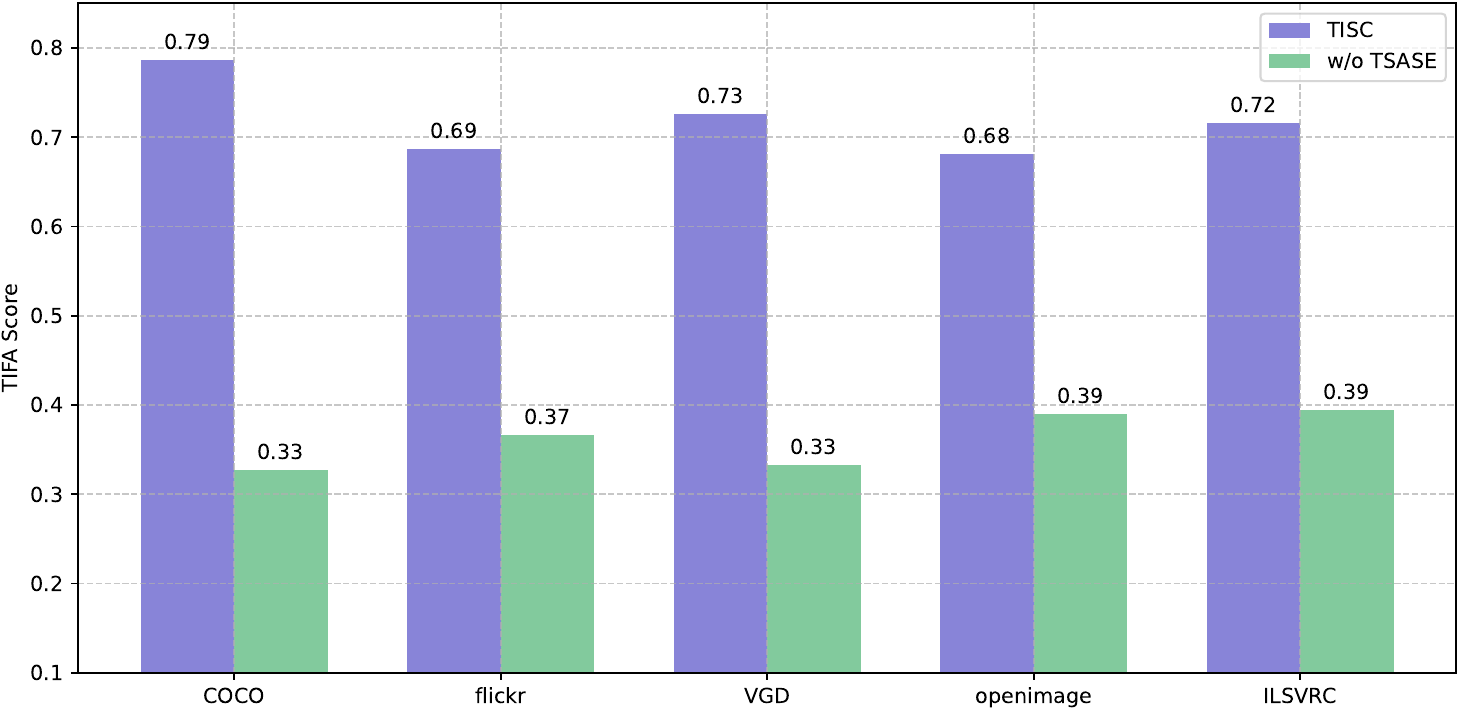}
    \caption{Average TIFA scores ($\overline{\mathrm{TIFA}}$) of TISC and w/o TSASE across five tested datasets.}
    \label{fig:avg_tifa}
\end{figure}

Next, we evaluate the accuracy of TSASE in extracting semantic description. Since TSASE aims not only to provide spatial positions, but also to generate more accurate object attributes, background information, and global scene descriptions, we employed the TIFA metric (Text-to-Image Faithfulness Assessment)~\cite{tifa} to quantify the consistency between the extracted semantic description and the original image. TIFA is an automated evaluation framework based on fine-grained Visual Question Answering (VQA) powered by an external MLLM, designed to measure the alignment between text-to-image generations and the ground-truth text descriptions. In our use of TIFA, only the original image and the transmitter-side text description are required, i.e., this evaluation does not involve the receiver-side image reconstruction process. Therefore, this experiment directly compares the semantic extraction accuracy of TSASE with the one-shot holistic description used in w/o TSASE.
\begin{figure*}[t]
    \centering
    \includegraphics[width=0.85\linewidth]{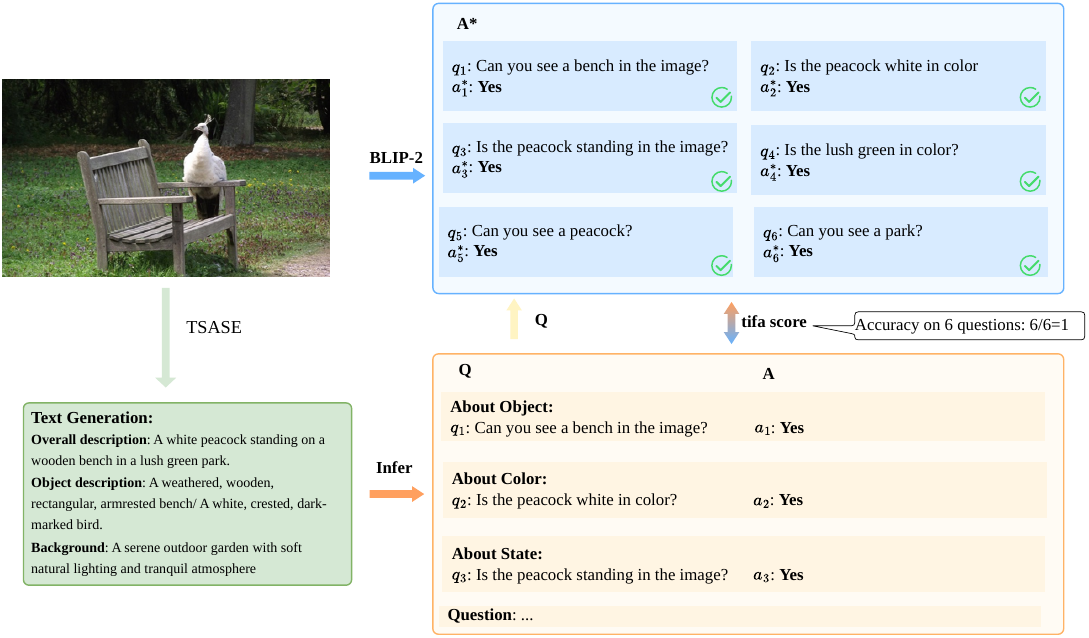}
    \caption{TIFA evaluation pipeline.}
    \label{fig:tifa_pipeline}
\end{figure*}

We introduce the TIFA evaluation pipeline with the illustration of Fig.~\ref{fig:tifa_pipeline}. TIFA applies the follow evaluation pipeline to both TISC and w/o TSASE:
\begin{enumerate}
    \item Given a text description $\bm{D}$, TIFA parses it into distinct semantic components (e.g., objects, color) for later processing.
    \item Observing components obtained in 1), TIFA asks a set of questions $Q=\{q_1,q_2,\ldots,q_n\}$ according to predefined rules and templates.
    \item a corresponding answer set $A=\{a_1,a_2,\ldots,a_n\}$ is derived from the description $\bm{D}$, where each question requires a binary (yes/no) answer (see the yellow part in Fig.~\ref{fig:tifa_pipeline}).
    \item To obtain the ground-truth answers that accurately reflects the semantics of the original image, $Q$ and $I$ are fed into a VQA model (such as BLIP-2\cite{blip2}) to produce standard answer set $A^*=\{a_1^*,a_2^*,\ldots,a_n^*\}$ (see the blue part in Fig.~\ref{fig:tifa_pipeline}).
\end{enumerate}

TIFA measures the alignment between text description $\bm{D}$ and image $I$ by comparing $A$ against $A^*$. TIFA score is defined as the accuracy over all questions, which is written as:
\begin{equation}
\mathrm{TIFA}(\bm{D},I)=\frac{1}{n}\sum_{i=1}^{n}\mathbb{I}(a_i=a_i^*),
\label{eq:tifa}
\end{equation}
where $\mathbb{I}(\cdot)$ is the indicator function that outputs $1$ if $a_i=a_i^*$ (otherwise, outputs $0$).

For a given dataset $\mathcal{T}$ with $N$ images $\{I_i\}_{i=1}^{N}$. The dataset-level average TIFA is defined as
\begin{equation}
\overline{\mathrm{TIFA}}=\frac{1}{N}\sum_{i=1}^{N}\mathrm{TIFA}(D_i,I_i),
\label{eq:avg_tifa}
\end{equation}

For TISC, we use the TSASE module to extract a comprehensive text description $\bm{D_{\mathrm{TSASE}}}$, and we have $\bm{D}=\bm{D_{\mathrm{TSASE}}}$ in the TIFA evaluation pipeline. For w/o TSASE, as introduced in the experimental set, we use GPT-4o \cite{gpt4o} to place the TSASE module, and we ask the model to directly generate an overall description of the image. Let us denote the text description generated by GPT-4o as $\bm{D_{\mathrm{4o}}}$.

Fig.~\ref{fig:avg_tifa} shows the average TIFA scores of TISC and w/o TSASE across five datasets, from which we see that TISC consistently achieves higher average TIFA score on all datasets.

\subsection{INO Parameter Study}
\label{subsec:InoParam}

This subsection further analyzes two key parameter settings in the INO mechanism. Unlike the TSASE ablation study, the performance of INO mainly depends on the search range of candidate initial noises and the evaluation criterion used to select among candidate reconstructions. Therefore, this subsection is organized into two experiments corresponding to the following two questions:

First, how should the number of candidate initial noises $m$ be determined to obtain a high-quality reconstruction with high probability?

Second, how should the weighting parameter $\alpha$ in the comprehensive similarity score $S_{\mathrm{comp}}$ be set to balance visual similarity and semantic similarity?

\textbf{Experiment 1: Selection of the Number of Noise Seeds}

We first address the selection of the candidate seed number $m$. Since different initial noise seeds may lead to different candidate reconstructions, the seed budget $m$ should ensure, with high probability, that the candidate set contains at least one high-quality reconstruction, while avoiding unnecessary candidate image generation. Our goal is to obtain, through rigorous statistical reasoning, a unified seed budget that is independent of any specific image. Otherwise, if the required number of seeds varied with the particular image, an additional method would be needed to identify the proper seed number for each image, introducing a cumbersome image-specific procedure.

To formalize the above seed budget selection problem, we introduce the following notation. Consider the $r$-th original image $I_r$. Suppose the INO module uses $m$ independent initial noise seeds to generate $m$ candidate reconstructions, denoted by $\{\hat{I}_{r,q}\}_{q=1}^{m}$. The corresponding comprehensive similarity scores are given by $\{S_{\mathrm{comp}}(I_r,\hat{I}_{r,1}),S_{\mathrm{comp}}(I_r,\hat{I}_{r,2}),\ldots,S_{\mathrm{comp}}(I_r,\hat{I}_{r,m})\}$. Here, $S_{\mathrm{comp}}(I_r,\hat{I}_{r,q})$ denotes the comprehensive similarity score between the original image $I_r$ and the corresponding candidate reconstruction $\hat{I}_{r,q}$ generated from the $q$-th seed. We further use $S_{\mathrm{comp}}^{(r)}$ to denote the random variable of the comprehensive similarity score induced by a randomly sampled seed for the $r$-th image. Since the INO module selects the reconstruction with the highest similarity score, the quality achieved under seed budget $m$ is determined by $\max_{q=1,\ldots,m}S_{\mathrm{comp}}(I_r,\hat{I}_{r,q})$.

Under this notation, the objective of seed budget selection is to choose the smallest possible $m$ such that the probability that at least one candidate in the set reaches a high-quality reconstruction threshold is no smaller than a prescribed confidence level $p$. Formally, we require
\begin{equation}
P\left(\max_{q=1,\ldots,m}S_{\mathrm{comp}}(I_r,\hat{I}_{r,q})\ge t_r(k)\right)\ge p.
\label{eq:seed_budget_requirement}
\end{equation}
Here, $t_r(k)$ denotes the image-dependent quality threshold for the $r$-th image, where $k$ controls how strict the high-quality criterion is (see Eq.~\eqref{eq:quality_threshold} for more details). Although $t_r(k)$ depends on the distribution parameters of each image, the resulting seed budget $m$ will be shown to depend only on predefined hyperparameter $k$ and $p$, rather than on the image index $r$. This property is important because it provides a uniform seed selection rule across all images. To evaluate the probability in Eq.~\eqref{eq:seed_budget_requirement} and solve for the required seed budget, we first need to characterize the distribution of the similarity score $S_{\mathrm{comp}}^{(r)}$.

To empirically characterize the distribution of $S_{\mathrm{comp}}^{(r)}$, we randomly select three images from the COCO image dataset and generate $500$ candidate reconstructions for each image using $500$ different initial noise seeds. For each generated candidate, we compute the comprehensive similarity score according to Eq.~\eqref{eq:scomp}, while keeping the semantic description, generation model, and inference configuration fixed. We then plot the histogram of the obtained comprehensive similarity scores and fit a Gaussian distribution to each histogram.

\begin{figure*}[t]
    \centering
    \includegraphics[width=\linewidth]{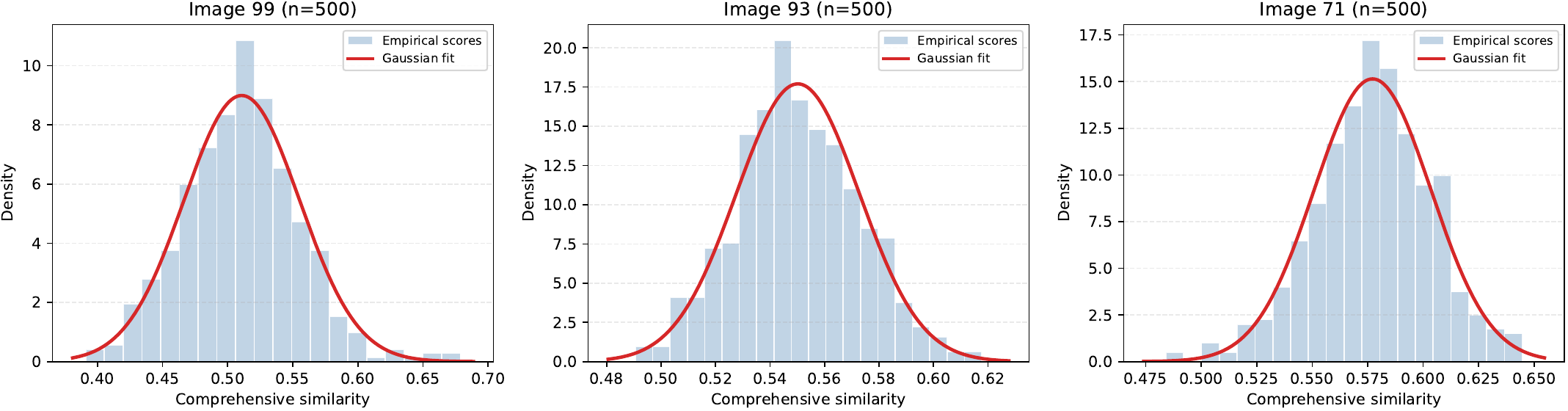}
    \caption{Empirical distribution of comprehensive similarity scores induced by random initial noise seeds. The three subfigures correspond to COCO image IDs 99, 93, and 71, respectively. For each image, we fix the semantic description, generation model, and inference configuration, and generate 500 candidate reconstructions using 500 different initial noise seeds. The comprehensive similarity score of each candidate is computed according to Eq.~\eqref{eq:scomp}. The blue histogram shows the empirical distribution of the 500 comprehensive similarity scores, where the y-axis denotes probability density rather than sample count, computed as the bin count divided by the product of the total number of samples and the bin width, so that the total area of the histogram is normalized to one. The x-axis represents the comprehensive similarity score. The red curve shows the Gaussian distribution fitted using the empirical mean and standard deviation of the comprehensive similarity scores.}
    \label{fig:score_distribution}
\end{figure*}

As shown in Fig.~\ref{fig:score_distribution}, for each fixed image, the comprehensive similarity scores produced by different initial noise seeds form a stable unimodal distribution. The fitted Gaussian curves are well aligned with the empirical histograms, indicating that the distribution of $S_{\mathrm{comp}}^{(r)}$ can be reasonably approximated by a Gaussian distribution. Therefore, for the $r$-th image, we model $S_{\mathrm{comp}}^{(r)}\sim\mathcal{N}(\mu_r,\sigma_r^2)$, where $\mu_r$ and $\sigma_r$ denote the mean and standard deviation of the comprehensive similarity scores under random seed sampling.

Under this Gaussian model, we define a high-quality candidate as a reconstruction whose comprehensive similarity score is significantly higher than the average seed outcome for the same image. Specifically, we set the high-quality threshold as
\begin{equation}
t_r(k)=\mu_r+k\sigma_r,
\label{eq:quality_threshold}
\end{equation}
where $k$ specifies how many standard deviations above the image-specific mean the high-quality threshold is set. A larger $k$ requires a candidate reconstruction to exceed the average similarity level by more standard deviations, and therefore corresponds to a stricter definition of high-quality reconstruction. In this paper, we use $k=1$ as the default setting, meaning that a candidate reconstruction is regarded as high-quality if its comprehensive similarity score is at least one standard deviation above the image-specific mean. For a Gaussian distribution, samples above $\mu_r+\sigma_r$ account for approximately the top $15.87\%$ of the distribution.

According to the Gaussian distribution, the probability that a single seed fails to reach this threshold is
\begin{equation}
P\left(S_{\mathrm{comp}}^{(r)}<t_r(k)\right)
=
P\left(\frac{S_{\mathrm{comp}}^{(r)}-\mu_r}{\sigma_r}<k\right)
=
\Phi(k),
\label{eq:single_seed_failure}
\end{equation}
where $\Phi(\cdot)$ is the cumulative distribution function of the standard normal distribution. It is important to note that this probability depends only on $k$, rather than on the image index $r$. Although different images may have different $\mu_r$ and $\sigma_r$, these image-dependent parameters are removed by standardization.

Based on the above derivation and the probability requirement in Eq.~\eqref{eq:seed_budget_requirement}, the required seed budget $m$ varies with the chosen parameter $k$ in $t_r(k)$. Therefore, we denote it as $m_k$ in the following derivation.

If $m_k$ independent candidate reconstructions are generated, the probability that none of them reaches the high-quality threshold is
\begin{equation}
P\left(\max_{q=1,\ldots,m_k}S_{\mathrm{comp}}(I_r,\hat{I}_{r,q})<t_r(k)\right)
=
\Phi(k)^{m_k}.
\label{eq:no_candidate_success}
\end{equation}
Thus, the probability that at least one candidate reaches the high-quality threshold is
\begin{equation}
P\left(\max_{q=1,\ldots,m_k}S_{\mathrm{comp}}(I_r,\hat{I}_{r,q})\ge t_r(k)\right)
=
1-\Phi(k)^{m_k}.
\label{eq:at_least_one_success}
\end{equation}
To ensure that this probability is no smaller than the prescribed confidence level $p$, we require
\begin{equation}
1-\Phi(k)^{m_k}\ge p.
\label{eq:confidence_constraint}
\end{equation}
Solving for $m_k$, we obtain
\begin{equation}
m_k\ge
\left\lceil
\frac{\ln(1-p)}{\ln(\Phi(k))}
\right\rceil.
\label{eq:seed_budget}
\end{equation}
Here, $\lceil\cdot\rceil$ denotes the ceiling operation, ensuring that the seed budget is the smallest integer satisfying the probability constraint.

In our default setting, we set $k=1$ and the target confidence level to $p=0.90$, which corresponds to a seed budget of $m_k=14$ under the above formulation. This means that the candidate set is expected to contain at least one high-quality reconstruction with $90\%$ probability. Therefore, in the subsequent experiments, we use $14$ initial noise seeds by default for candidate reconstruction generation and selection.

Beyond the specific value of $m_k=14$ used in the subsequent experiments, the above procedure can also be regarded as a practical calibration example for configuring INO. Rather than claiming $m_k=14$ as a universal seed budget, our goal is to demonstrate a reproducible procedure for selecting the seed-search budget under a given deployment setting. In practice, the appropriate seed budget may depend on the target image domain, the adopted generative model, the similarity metric, and the available transmitter-side computational budget. A system designer can follow the same procedure used in this paper: sample candidate seeds under the target deployment setting, characterize the resulting reconstruction-quality distribution, and select $m_k$ according to the desired confidence level and quality threshold.

\textbf{Experiment 2: Validation of the Weighting Parameter $\alpha$}

After determining the candidate seed number as $m=14$ in Experiment~1, we further study how to set the weighting parameter $\alpha$ between visual similarity and semantic similarity in $S_{\mathrm{comp}}$. Since candidate selection is directly determined by $S_{\mathrm{comp}}$, an inappropriate $\alpha$ may bias the selection toward either visually similar but semantically incomplete reconstructions or semantically similar but visually less faithful reconstructions.

To investigate this effect, we consider five weighting configurations for $S_{\mathrm{comp}}$. For each original image, all configurations select one reconstruction from the same candidate set, and they differ only in the value of $\alpha$:
\begin{itemize}
    \item Configuration A: $\alpha=1$, using visual similarity only;
    \item Configuration B: $\alpha=0.8$;
    \item Configuration C: $\alpha=0.5$, assigning equal weights to visual and semantic similarities;
    \item Configuration D: $\alpha=0.2$;
    \item Configuration E: $\alpha=0$, using semantic similarity only.
\end{itemize}

Here, Configurations A and E correspond to visual-only and semantic-only selection, respectively. Configuration C assigns equal weights to the two similarity terms, while Configurations B and D place more emphasis on visual similarity and semantic similarity, respectively. By comparing the reconstructions selected under these configurations, we can analyze the individual and combined effects of visual and semantic similarities on candidate reconstruction selection.

We next describe how different weighting configurations are evaluated on a test dataset. We first introduce the notation used. Given a test dataset $\mathcal{T}=\{I_r\}_{r=1}^{N_{\mathrm{test}}}$, for the $r$-th original image $I_r$, INO first generates a candidate reconstruction set using $m=14$ different random seeds, denoted by $\Omega=\{\hat{I}_{r,q}\}_{q=1}^{m}$.

Let $L=\{A,B,C,D,E\}$ denote the set of five weighting configurations. For a given configuration $l\in L$, we compute $S_{\mathrm{comp}}$ for each reconstruction in the candidate set using the corresponding value of $\alpha$, and select the reconstruction with the highest score as the output of this configuration:
\begin{equation}
\hat{I}_r^{(l)}
=
\arg\max_{\hat{I}_{r,q}\in\Omega}
S_{\mathrm{comp}}^{(l)}(I_r,\hat{I}_{r,q}),
\label{eq:config_output}
\end{equation}
where $S_{\mathrm{comp}}^{(l)}$ denotes the comprehensive similarity score computed under the value of $\alpha$ associated with configuration $l$. Therefore, for each original image $I_r$, the five weighting configurations produce five configuration outputs, denoted by $\{\hat{I}_r^{(l)}\}_{l\in L}$.

To determine which weighting configuration is more suitable, we introduce an external evaluator (i.e., either a human annotator or a human-validated MLLM). The evaluator selects, from the five configuration outputs $\{\hat{I}_r^{(l)}\}_{l\in L}$, the reconstruction that is most similar to the original image by jointly considering visual appearance and semantic content. Let $\hat{I}_r^*$ denote the reconstruction selected by the external evaluator for the $r$-th original image. Since different weighting configurations may select the same reconstruction from the same candidate pool $\Omega$, a configuration is considered to have selected the best reconstruction for the $r$-th sample if its output satisfies $\hat{I}_r^{(l)}=\hat{I}_r^*$.

For a given weighting configuration $l$, its best-reconstruction selection rate over the test dataset is defined as
\begin{equation}
R_l=
\frac{1}{N_{\mathrm{test}}}
\sum_{r=1}^{N_{\mathrm{test}}}
\mathbb{I}\left(\hat{I}_r^{(l)}=\hat{I}_r^*\right),
\label{eq:selection_rate}
\end{equation}
where $\mathbb{I}(\cdot)$ is the indicator function. $R_l$ represents the proportion of test samples for which configuration $l$ successfully selects the reconstruction considered closest to the original image by the external evaluator. Therefore, $R_l$ is used as the dataset-level metric for comparing different weighting configurations in INO. The remaining issue is how to determine a suitable external evaluator for obtaining $\hat{I}_r^*$.

Human evaluation can directly reflect human perceptual preferences and can therefore serve as the external evaluator described above. However, since this paper needs to compare different weighting configurations over the full test sets of multiple datasets, relying entirely on human annotation would incur a high annotation cost. Therefore, we employ GPT-4o as the MLLM-based external evaluator to evaluate different weighting configurations at the dataset scale. To ensure that the MLLM judgments can reasonably reflect human preferences, we first construct a small-scale human validation dataset before applying the MLLM to the full test datasets.

Specifically, we randomly sample $20$ original images from each dataset, resulting in $100$ original images in total. For each original image $I_r$ in the validation set, we obtain five configuration outputs $\{\hat{I}_r^{(l)}\}_{l\in L}$ following the procedure described above. Both human evaluation and MLLM-based evaluation are performed on these five outputs, where the evaluator selects the reconstruction that is overall closest to the original image.

Note that, during this evaluation stage, revealing the weighting configuration, the value of \(\alpha\), the associated similarity score, or the method name for a given output may introduce subjective bias unrelated to the image content. To avoid this potential bias, we use an anonymized candidate presentation protocol for both human and MLLM-based evaluations. Specifically, the five configuration outputs are presented to the evaluator using only anonymous labels A, B, C, D, and E, without revealing their corresponding weighting configurations, \(\alpha\) values, \(S_{\mathrm{comp}}\) scores, or method names.

Based on this anonymized presentation protocol, in the human evaluation stage, for each original image $I_r$, a human annotator is asked to select, from $\{\hat{I}_r^{(l)}\}_{l\in L}$, the reconstruction that is most similar to the original image by considering both semantic content and visual appearance. The human-selected reconstruction is denoted by $\hat{I}_r^{*,(\mathrm{human})}$.

\begin{figure}[t]
    \centering
    \includegraphics[width=\linewidth]{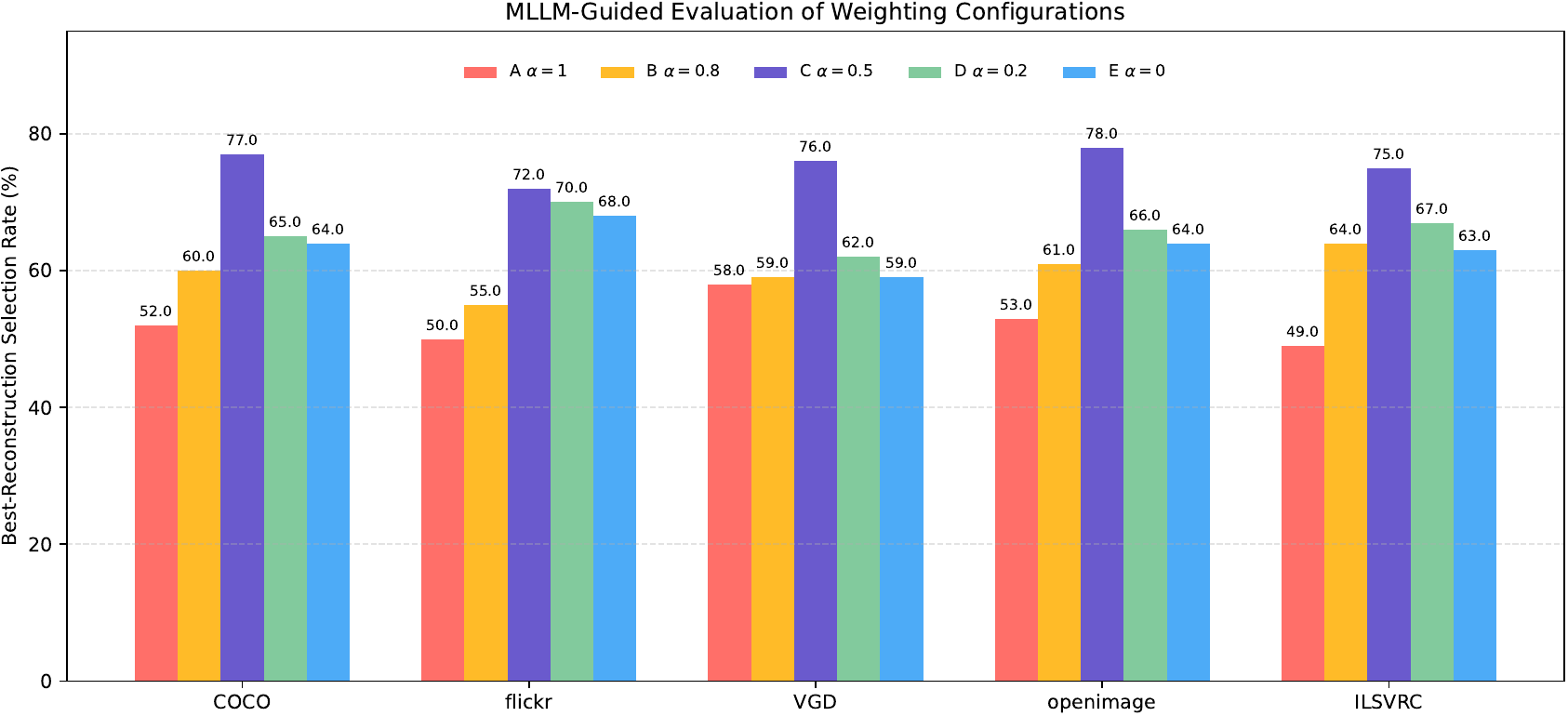}
    \caption{MLLM-guided evaluation of different weighting configurations.}
    \label{fig:weighting_configuration}
\end{figure}

For MLLM-based annotation, each original image and its five corresponding reconstructed candidates are jointly provided to the MLLM. The MLLM then selects the candidate that is most similar to the original image. The prompt used for this evaluation is as follows:
\begin{quote}
You are given one reference image and five reconstructed candidate images. Please select the reconstructed candidate that is most similar to the reference image. When making the selection, consider both visual appearance and semantic content in an overall manner. You must choose exactly one candidate from A, B, C, D, and E. Return only a JSON object in the following format:

\texttt{\{"winner": "A"\}}
\end{quote}
This prompt asks the MLLM to evaluate the reconstructed candidates from both visual and semantic perspectives. To facilitate automatic parsing, we require the MLLM to return only a JSON object containing the field \texttt{winner}. The MLLM-selected reconstruction is denoted by $\hat{I}_r^{*,(\mathrm{MLLM})}$.

Based on the above, the human-MLLM agreement rate on the validation set is defined as
\begin{equation}
A_{\mathrm{agree}}=
\frac{1}{N_{\mathrm{val}}}
\sum_{r=1}^{N_{\mathrm{val}}}
\mathbb{I}\left(\hat{I}_r^{*,(\mathrm{MLLM})}=\hat{I}_r^{*,(\mathrm{human})}\right),
\label{eq:agreement_rate}
\end{equation}
where $N_{\mathrm{val}}=100$ denotes the number of original images in the validation set, and $\mathbb{I}(\cdot)$ is the indicator function. In our validation experiment, the MLLM achieves a $93\%$ agreement rate with human selections, indicating that its judgments are well aligned with human perceptual preferences in this evaluation setting.

Therefore, we further adopt the MLLM as the external evaluator to assess different weighting configurations on the full test datasets. Based on the MLLM selections, we compute the best-reconstruction selection rate $R_l$ for each weighting configuration $l$ on each tested dataset (see Eq.~\eqref{eq:selection_rate} for more details).

As shown in Fig.~\ref{fig:weighting_configuration}, Configuration C achieves the highest best-reconstruction selection rate across all datasets. This result indicates that assigning equal weights to visual similarity and semantic similarity provides a better balance between perceptual consistency and semantic faithfulness. Therefore, we set $\alpha=0.5$ in $S_{\mathrm{comp}}$ as the default weighting configuration for INO.

\subsection{End-to-End Robustness Analysis in Wireless Channels}

In the preceding experiments, we mainly evaluated TISC as a text-driven semantic source coding and decoding framework under the reliable-delivery assumption. To further examine its robustness over wireless channels, we extend TISC into an end-to-end wireless communication system. Specifically, the text semantic description generated at the transmitter and the noise seed selected by INO are first converted into a bit stream using UTF-8 encoding. The bit stream is then protected by an LDPC channel code with rate $3/8$, modulated using BPSK, and transmitted over an AWGN channel. At the receiver, the received signal is demodulated and channel-decoded to recover the text semantic information and the noise seed, which are then used for image reconstruction. Under this setting, we consider AWGN channels with different SNRs to evaluate the end-to-end noise robustness of TISC.

To compare TISC with conventional image transmission and existing semantic communication schemes over wireless channels, we introduce the following baselines:

\textbf{JPEG baseline:} JPEG is a commonly used lossy image source coding standard \cite{jpeg}. In this baseline, each image is first compressed into a JPEG bitstream using a JPEG encoder with quality factor $Q=1$ \footnote{Since the implementation of the JPEG quality factor may vary across different encoders, we use the JPEG encoder provided by the Pillow library~\cite{pillow_jpeg} with ``quality=1'' for reproducibility.}. The resulting JPEG bitstream is then protected by the same LDPC channel code with rate $3/8$, modulated using BPSK, and transmitted over the same AWGN channel. At the receiver, if a valid image cannot be reconstructed after standard JPEG decoding and image recovery procedures, the sample is treated as a reconstruction failure. For such failed samples, a zero image similarity score is assigned.

\textbf{DeepJSCC:} DeepJSCC is an end-to-end joint source-channel coding framework for image semantic transmission \cite{Deepjscc}. It uses neural networks to directly map the input image into continuous channel symbols for transmission and reconstructs the image from the received symbols at the receiver. Therefore, the source coding, channel coding, and modulation-related operations are implicitly modeled within the end-to-end learned system.

The experiment is conducted on 100 images randomly sampled from the COCO dataset. For each method, we compute the average comprehensive similarity score (between the transmitted image and the recovered one) over this dataset under different SNR conditions, where the score is defined in Eq.~\ref{eq:scomp}.

\begin{figure}[t]
    \centering
    \includegraphics[width=\linewidth]{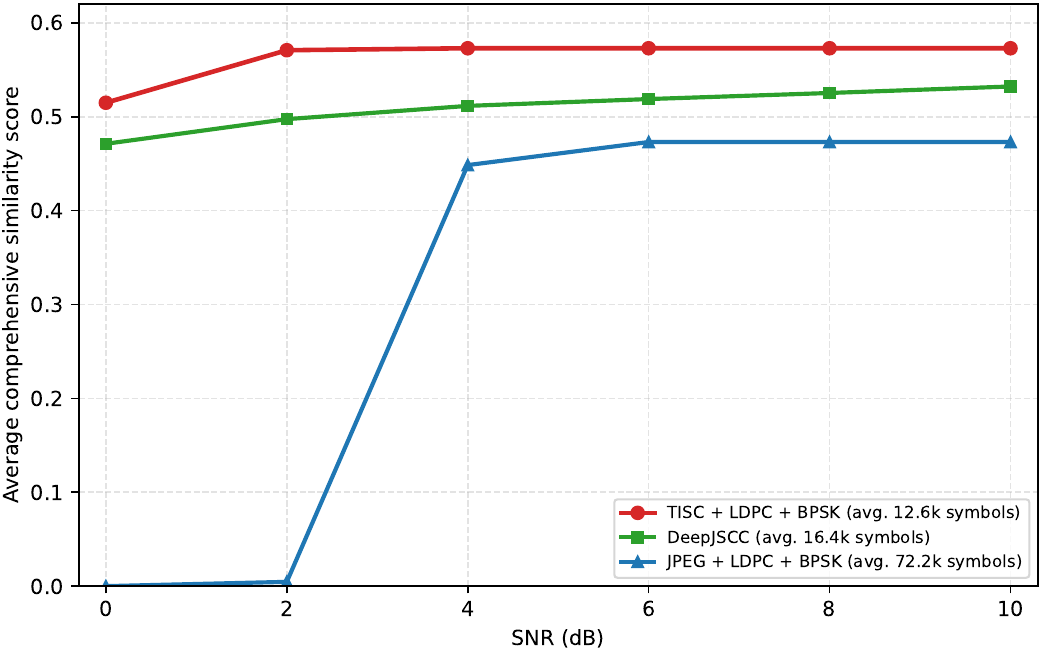}
    \caption{End-to-end noise robustness comparison over an AWGN channel. To characterize the channel resource usage of different schemes, we report the average number of channel symbols used per image. For TISC and the JPEG baseline, this number is counted from the output symbols after BPSK modulation. For DeepJSCC, since its end-to-end model implicitly includes source coding, channel coding, and modulation-related operations, we directly count the number of channel symbols produced by its encoder. Under this experimental setting, TISC, the JPEG baseline, and DeepJSCC use approximately 12.6k, 72.2k, and 16.4k channel symbols per image on average, respectively. This indicates that the performance advantage of TISC is not obtained by using more channel resources.}
    \label{fig:end_to_end_robustness}
\end{figure}

As shown in Fig.~\ref{fig:end_to_end_robustness}, TISC achieves higher comprehensive similarity scores than both baselines across all SNR values tested. In addition to the superior performance, TISC uses fewer channel symbols than DeepJSCC and substantially fewer channel symbols than the JPEG baseline (see the caption of Fig.~\ref{fig:end_to_end_robustness} for statistic details).

At the extremely low SNR of 0~dB, 32 out of the 100 image transmissions cannot fully recover the transmitted text. Yet, our TISC receiver can still generate a highly related image based on the partly corrupted semantic information in text (with an average similarity score of 0.5149), while DeepJSCC and the standard JPEG baseline perform worse in such conditions. This result indicates that the text semantic representation in TISC provides a certain degree of error tolerance: \textbf{\textit{even when part of the received text is corrupted, the receiver may still exploit the remaining semantic information to generate an image that is semantically related to the original one}}.

When the SNR increases to 2~dB, the text semantic information of only 4 out of 100 images is not exactly recovered, and TISC achieves an average similarity score of 0.5708, which is already close to its error-free performance of 0.5729. From 4~dB onward, both the text semantic information and the selected noise seed are fully recovered for all test images, and thus the TISC performance remains stable as the SNR further increases. In contrast, the JPEG bitstream is more sensitive to residual bit errors and may fail to decode under low-SNR conditions. DeepJSCC exhibits a smoother performance variation with SNR, but its reconstruction similarity remains lower than that of TISC in this setting.

%% file: sections/05_conclusion.tex
\section{Conclusion}
This paper proposed TISC, a text-driven image semantic communication framework for semantically faithful image reconstruction. Two major techniques, namely Tree-Structured Attribute Semantic Extraction (TSASE) and Initial Noise Optimization (INO), have been developed in our TISC framework: TSASE extracts a structured semantic description to reduce information loss and distortion, while INO selects the most favorable initial noise seed at the transmitter to improve the visual and semantic consistency of receiver-side reconstruction. We have conducted massive experiments to validate the effectiveness of the two techniques proposed. Our results on multiple datasets show that TSASE substantially improves both object-position recovery and semantic description faithfulness over the without-TSASE ablation. Specifically, across all datasets, TISC improves object-position alignment by at least 42 percentage points and achieves text-image faithfulness scores above 0.68, compared with only 0.33–0.39 for this ablation. Furthermore, parameter study on INO determines reasonable settings for the number of candidate noises and the weighting between visual and semantic similarities in the comprehensive similarity score, providing practical guidance for configuring INO in the proposed system. These results validate the effectiveness and practical feasibility of the proposed framework for deployment in future wireless networks.

%% file: sections/appendix_a.tex
\section{Experimental Evidence Supporting the Motivation}
\label{sec:appendix}

This appendix provides a motivating experiment to illustrate the bandwidth-saving potential of text-driven image semantic communication. We compare JPEG-based image transmission with the proposed text-driven semantic transmission scheme, TISC, where JPEG~\cite{jpeg} is a classical lossy image compression standard. The comparison is conducted in terms of the number of compressed bits to be transmitted and the comprehensive similarity score between the original and reconstructed images. The comprehensive similarity score jointly measures visual and semantic similarities between the two images (see Eq.~\ref{eq:scomp} for more details).

\begin{figure}[t]
    \centering
    \includegraphics[width=\linewidth]{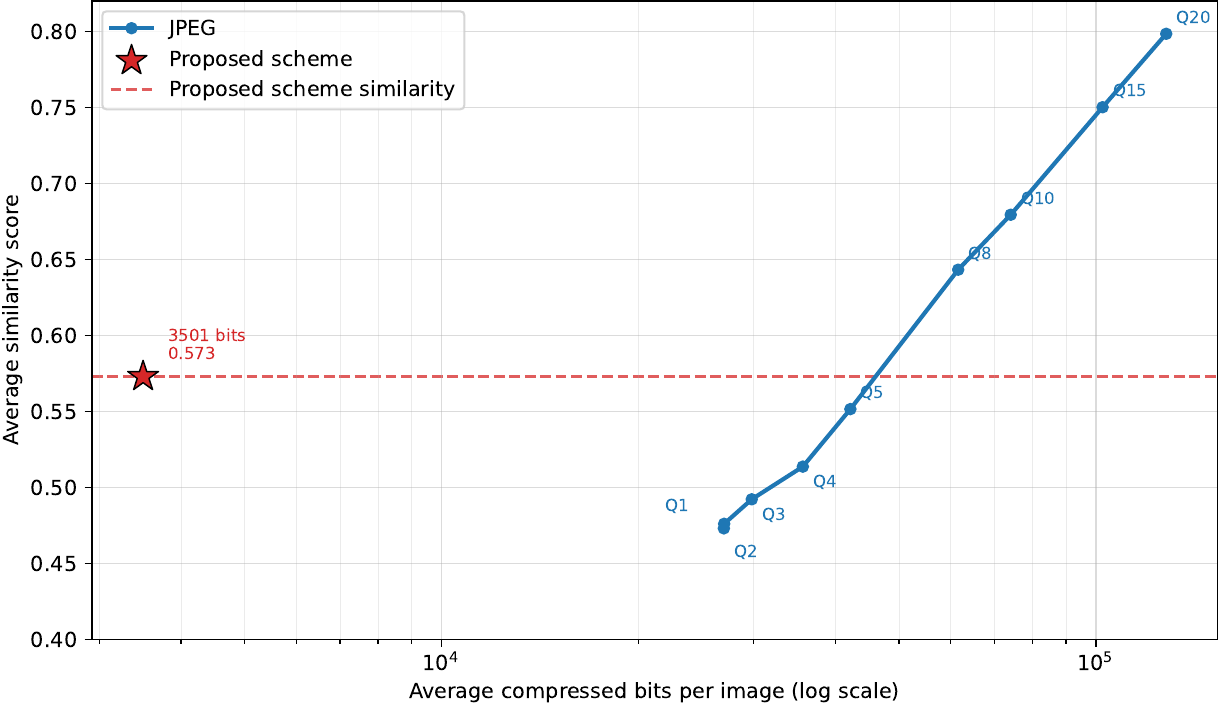}
    \caption{Comparison between JPEG-based image transmission and the proposed text-driven semantic transmission scheme in terms of compressed bits and the comprehensive similarity score. The x-axis reports the average number of compressed bits per image, and the y-axis reports the dataset-level average comprehensive similarity score computed according to Eq.~\ref{eq:scomp}.}
    \label{fig:appendix_jpeg}
\end{figure}

The experiment is conducted on 100 images sampled from the COCO dataset. For each test image, we apply both JPEG and TISC to obtain the corresponding reconstructed images, and then compute the comprehensive similarity score between each reconstruction and the original image. For JPEG-based transmission, we vary the JPEG quality factor to obtain reconstructions under different compression levels. For each quality factor, we compute the average number of compressed bits and the average comprehensive similarity score over the 100 images, thereby forming a curve between bit cost and reconstruction similarity. For TISC, the image is first compressed into text semantic information to be transmitted, and the required number of bits is computed from the UTF-8 encoding length of this text semantic information. The average bit cost and average comprehensive similarity score are also computed over the same 100 images. To focus on the compression efficiency of different source representations, we only count the bits after source compression/encoding, without including channel coding overhead, and assume reliable information delivery.

As shown in Fig.~\ref{fig:appendix_jpeg}, at the reconstruction similarity level achieved by TISC, JPEG requires more than 20 times as many compressed bits to reach a comparable similarity level. This result provides empirical evidence for the motivation of text-driven image semantic communication: by transmitting compact text semantic information instead of image pixels or dense visual features, the communication system can substantially reduce the transmitted bit cost while preserving semantic similarity in the reconstructed image.